\pdfoutput=1
\documentclass{llncs}
\usepackage{llncsdoc}    
\usepackage[utf8]{inputenc}
\begin{document}
\newcounter{save}\setcounter{save}{\value{section}}
{\def\addtocontents#1#2{}%
\def\addcontentsline#1#2#3{}%
\def\markboth#1#2{}%
\title{An English-Hindi Code-Mixed Corpus: \\Stance Annotation and Baseline System}

\author{\textbf{Sahil Swami}, \textbf{Ankush Khandelwal}, \textbf{Vinay Singh}, \textbf{Syed Sarfaraz Akhtar} \and \textbf{Manish Shrivastava}}
\institute{Language Technologies Research Centre, International Institute of Information Technology, Hyderabad}

\maketitle
\begin{abstract}
Social media has become one of the main channels for people to communicate and share their views with the society. We can often detect from these views whether the person is in favor, against or neutral towards a given topic. These opinions from social media are very useful for various companies. We present a new dataset that consists of 3545 English-Hindi code-mixed tweets with opinion towards Demonetisation that was implemented in India in 2016 which was followed by a large countrywide debate. We present a baseline supervised classification system for stance detection developed using the same dataset that uses various machine learning techniques to achieve an accuracy of 58.7\% on 10-fold cross validation.
\end{abstract}
\section{Introduction}
In recent times, social media platforms such as Facebook, Twitter, Linkedin have gained a lot of popularity. They offer people the medium to connect with friends, family and colleagues, and express their opinions freely and. This leads to people expressing themselves a lot on these platforms. We can find opinions on almost any topic may it be sports, politics or movies. The researchers call this kind of data, the \textit{Big Data}, characterized by ``3V" which stands for \textit{Volume, Variety} and \textit{Velocity}. Some also refer to it as ``5V" i.e. for \textit{Value} and \textit{Veracity} \cite{1}. Analysis of these opinions in big data is referred to as opinion mining or stance detection. 

Stance detection is the task of automatically determining from text whether the author is in favor or against or is neutral towards a target. The target in this paper is `Notebandi' or `Demonetisation' which was implemented in India on 8\textsuperscript{th} November, 2016 in which currency in the denominations of 500 and 1000 was declared invalid. The government claimed that this decision was taken to eliminate the use of counterfeit cash used to fund illegal activities and terrorism. People all over India had different reactions to this event and many of them used twitter to express their views. Consider the following tweet: `\textit{Demonetisation has caused a lot of problems for everyone}'. We can say that the author if this tweet is most likely to be against the target demonetisation.

There have been several experiments in the field of opinion mining on social media and online texts \cite{2},\cite{3}. Opinion mining can provide a lot of information about the texts present in social media and can benefit many other tasks such as information retrieval, text summarization, etc.

Code-mixing is the conversion of one language to another within the same utterance or in the same oral or written text \cite{4}. This phenomenon is generally intra-sentential. and is common in multilingual societies. With 41\% of the Indian population speaking Hindi and English as the lingua franca of India, English-Hindi is the most commonly used code-mixed language pair on social media. An example of an English-Hindi code-mixed sentence is: `\textit{Dear sir Lagta hai bina tayari ka notebandi hua hai 2000 ka note ka size kam nahi karna chahiye}'. This sentence contains words in English such as `Dear', `sir' and words in Hindi such as `hai', `bina', etc. which are transliterated to English.

Several code-mixed datasets have been created \cite{5},\cite{6} for different NLP tasks but no opinion mining experiment has been performed on English-Hindi code-mixed data. Therefore, we aim to provide an English-Hindi code-mixed dataset and perform an experiment of opinion mining on it.

The main contribution of this paper is a resource of English-Hindi code-mixed tweets on `Notebandi' or `Demonetisation' with tweet level annotation for stance towards this target and token level language annotation that can be used to develop and evaluate the performance of stance detection and language identification techniques on code-mixed corpus. We also present a baseline classification system for stance detection on the same corpus.

Both the dataset and classification system are available online\footnote{https://github.com/sahilswami96/StanceDetection\_CodeMixed}.
\section{Dataset}

\subsection{Data Collection}
We collect tweets related to the Demonetisation that was implemented in India in 2016. We use Twitter Scraper API to collect tweets using the the keywords ’notebandi’ and ‘demonetisation’ over a period of 6 months after Demonetisation was implemented. All the tweets that are written exclusively in English or Hindi are eliminated and code-mixed tweets are selected manually. Each tweet is collected in json format after which the content of the tweet and the tweet id are extracted from it. A total of 3545 English-Hindi code-mixed tweets are collected.
\subsection{Data Processing and Annotation}
The tweets are annotated by a group of native Hindi speakers who are also fluent in English. Each tweet is annotated for stance towards demonetisation. Tweets are then tokenized for language annotation after which the tokenization and language tags are manually reviewed to resolve any errors. The inter- annotator agreement i.e. Cohen’s Kappa on the annotations for stance \cite{15} turned out to be 0.82.
\subsubsection{Stance Annotation}
Each of the tweets is manually annotated with one of the following stance tags: `FAVOR’, `AGAINST’ and `NONE’. Some hashtags and keywords, such as \#IAmWithModi, \#ByeByeBlackMoney and `samarthan’ are direct indicators that the author is in favor of demonetisation. Similarly hashtags such as \#StopDemonetisation,  \#NoteNahiPMBadlo and \#ModiSurgicalStrikeOnCommonMan are clear indicators that the author is against demonetisation. Examples of tweets (with translation in English) with different stances towards the target are:\\

Target: Demonetisation\\

\textit{Tweet: @narendramodi thanks for notebandi hum aap ke saath hai}

\textit{Translation: @narendramodi thanks for notebandi we are with you}

Stance: FAVOR\\

\textit{Tweet: @PMOIndia Chalo Modi ji apne Deshwasi sang majak kar liya, ab log bahut paresan hai 500/1000 pr rahem kr Notebandi wapos lo}

\textit{Translation: @PMOIndia Modi ji you played a prank with the people of your country, people are really hassled. Show mercy on 500/1000 and take demonetization back}

Stance: AGAINST\\

\textit{Tweet: Neta samajh. Nahi pa rahe hai ki notebandi par hindu muslim rajniti kaise kare , ye hi hai sabka sath sabka vikas}

\textit{Translation: Ministers are confused on how to do hindu muslim politics on demonetization, this is everyone's unity everyone's progress}

Stance: NONE
\subsubsection{Tokenization and Language Annotation}
Several experiments have been performed for language identification \cite{7},\cite{8},\cite{9} on monolingual and code-mixed texts which motivates the task of token level language annotation in the presented corpus.

The text written on twitter by users is sometimes a lot different from normal texts found in documents. It is a common trend to use multiple punctuations and white spaces such as “...”, “,,,”, “!!!”, etc. It is also common to use multiple mentions, hashtags and URLs in tweet. We tokenize the tweets after taking this information into account and by using white spaces as delimiters. Tokenization is manually verified by multiple people proficient in both English and Hindi to correct any mistakes.

Each token is then annotated with one of the language tags: ‘en’, ‘hi’, ‘rest’. ‘En’ refers to English and is assigned to English words such as `happy', `today', etc. ‘hi’ refers to Hindi and is assigned to Hindi words transliterated in English such as `nahi' (no), `samajh' (understand). A token is annotated with ‘rest’ when it is a named entity, punctuation, hashtag, URL or a mention, etc. Initially the tokens are automatically annotated with language tags using online available dictionaries such as `Enchant' and the `rest' tag is assigned by identifying hashtags, URLs, mentions and emoticons. We also create a list of popular named entities related to Demonetization to annotate named entities. Then each tag is manually verified to correct any wrong annotation. Table 1. shows an example of a language annotated tweet.
\\
\begin{table}[h!]
\begin{center}
\begin{tabular}{|c|c|}

\hline
\textbf{Token} & \textbf{Language}\\
\hline
\#Notebandi&hi\\
\hline
ka&hi\\
\hline
niyam&hi\\
\hline
:&rest\\
\hline
khata&hi\\
\hline
nahi&hi\\
\hline
hai&hi\\
\hline
to&hi\\
\hline
khulwao&hi\\
\hline
.&rest\\
\hline
Aam&hi\\
\hline
aadmi&hi\\
\hline
:&rest\\
\hline
khulwa&hi\\
\hline
to&hi\\
\hline
lun&hi\\
\hline
.&rest\\
\hline
Par&hi\\
\hline
bhai&hi\\
\hline
bank&en\\
\hline
main&hi\\
\hline
ghusub&hi\\
\hline
Kasey&hi\\
\hline
?&rest\\
\hline
\end{tabular}
\caption{A tweet with token level language annotation}
 \end{center}
\end{table}
\subsection{Dataset analysis}
The dataset consists of 3545  English-Hindi code-mixed tweets where each of them is annotated with stance towards Demonetisation. Each tweet is tokenized and each token is annotated with a language tag. The dataset has 964 tweets in favor, 647 tweets against and 1934 tweets that have no stance towards the target. The average length of a tweet is 21.3 tokens per tweet. There are an average of 16.3, 2.0 and 3.0 `hi', `en' and `rest' tokens respectively per tweet. Table 2. shows corpus level statistics whereas Table 3. shows tweet level statistics. This corpus can be used for developing and evaluating opinion mining and language identification techniques.

\begin{table}[h!]
\centering
\begin{tabular}{|c|c|}
\hline
\textbf{Category} & \textbf{Number of tweets}\\
\hline
Total tweets&3545 \\
\hline
Tweets in favor&964 \\
\hline
Tweets against&647 \\
\hline
Neutral tweets&1934  \\
 \hline
 \end{tabular}
\caption{Corpus level statistics}
\end{table}
\begin{table}[h!]
\centering
\begin{tabular}{|c|c|}
\hline
\textbf{Category} & \textbf{Number of tokens}\\
\hline
Avg. tokens&21.3 \\
\hline
Avg. en tokens&2.0 \\
\hline
Avg. hi tokens&16.3 \\
\hline
Avg. rest tokens&3.0  \\
 \hline
 \end{tabular}
\caption{Tweet level statistics}
\end{table}
\subsection{Dataset structure}
The corpus is structured into three files. The first file contains a tweet id followed by the corresponding tweet text and a blank line and so on. The second file consists of tweet ids followed by language annotated tweets. The third file has the stance for each tweet. Each tweet id is followed by one of the stance tags and a blank line.

\section{Stance detection system}
We present a baseline system for stance detection for English-Hindi code-mixed tweets which uses various character and word level features. We run various machine learning models over these features for stance detection.

\subsection{Preprocessing}
URLs, mentions and stop words are removed from the tweets. Hashtags are extracted for each tweet and as it is a general trend to use camel case format while writing hashtags, we remove the `\#' from the hashtags and use an approach \cite{14} for hashtag decomposition to extract all the words from the hashtag. For example \#IAmWithModi can be decomposed into four separate words i.e. `I', `Am', `With' and `Modi'. Each of these words is then treated as a separate token.
\subsection{Features}
\subsubsection{Character N-grams}
Character n-gram refers to presence or absence of contiguous sequence of n characters in the tweet. It can be seen from previous works \cite{3},\cite{11} that character level features have a significant effect on stance detection.
We extract character n-grams for all values of n between 1 and 3. Including all the n-grams increases the size of feature vector enormously. Therefore, we consider only those n-grams in our feature vector which occur at least 8 times in the dataset. This reduces the size of feature vector significantly and also removes noisy n-grams.
\subsubsection{Word N-grams}
Word n-gram refers to presence or absence of contiguous sequence of n words or tokens in the tweet. Word n-grams have proven to be important features for stance detection in previous studies \cite{2},\cite{3},\cite{11}. We extract word n-grams for all values of n between 1 and 5. We include only those n-grams in our feature vector which occur at least 10 times in the dataset.
\subsubsection{Stance Indicative Tokens}
This feature refers to the presence or absence of stance indicative tokens. We use a variation of the approach to find stance indicative hashtags \cite{3} and extract stance indicative tokens for each language label. We calculate a score for each token for stance where score is defined as :\\				
\[Score(token) =
max_{label\in{Stance-Set}} \frac{freq(token,stance\_label)}{freq(token)}\]

where Stance-Set = \{FAVOR, AGAINST, NONE\}.	\\			

We consider only those tokens as features for stance indication which have a score $\geq$ 0.6 and occur at least five times in the dataset. We find such tokens for each of the language tags and consider them in the feature vector. The threshold value for scores and number of occurrences has been decided after empirical fine tuning. 
\subsection{Feature Selection}
Previous studies \cite{10},\cite{11} have shown that feature selection algorithms improve efficiency and accuracy of classification systems. We use chi square feature selection algorithm which uses chi-squared statistic to evaluate individual feature with respect to each class. This algorithm was used in order to extract the best features and reduce the size of feature vectors to 500.
\subsection{Classification Approach}
We compare various machine learning models using the same features for stance detection.
Three classification techniques have been used for this experiment:\\
1. Support Vector Machine with Radial Basis Function kernel\\
2. Random Forest classifier\\
3. Linear support vector machine\\
We use the scikit-learn implementation of these methods.					
After pre-processing the dataset and extracting all the desired features, we run the above mentioned techniques and perform 10-fold cross validation. 10-fold cross validation is run for each of the individual features separately to observe the effect of each feature on classification.
\subsection{Results}
Table 4. shows the accuracy achieved when considering a single feature at a time as well as considering all at the same time for each of the machine learning techniques. It can be observed that Support Vector Machine with Radial Basis Function kernel performs the best. It can also be observed that all of the three features have nearly the same effect on classification.

\begin{table}[h!]
\centering
\begin{tabular}{|c|c|c|c|}
\hline
\textbf{Features} & \textbf{RBF Kernel SVM} & \textbf{Random Forest} & \textbf{Linear SVM}\\
\hline
Character n-grams&55.4&54.6&51.2 \\
\hline
Word n-grams&54.6&54.2&54.4 \\
\hline
Stance indicative tokens&54.7&54.5&54.8 \\
\hline
All features&58.7&54.7&56.6  \\
 \hline
 \end{tabular}
\caption{Accuracies for all the three classifiers}
\end{table}

\section{Conclusion}
Opinion mining is being used in various applications today. With the abundance of on-going research in this field, it has become on of the most important tasks on big data.

We started by presenting the first English-Hindi code-mixed dataset collected from twitter for stance identification towards Demonetisation. We explained the methods used for annotating each tweet with stance towards the target and for annotating each token with a language tag. We then presented a framework for stance detection developed using the same dataset which uses three different machine learning techniques. These techniques are then evaluated by running 10-fold cross validation.

\section{Future work}
This dataset can further be improved by normalizing each token which will further improve the performance of the classification system. This dataset can also be used for developing systems for automatic language identification in code-mixed texts.

Similar datasets can be created with other language pairs and with multiple targets in the same dataset.

There is also a huge scope of improvement in the system developed for stance detection by exploring other features such as POS tags and using various other machine learning techniques such as neural networks.


\end{document}